\documentclass[11pt,a4paper]{article}
\usepackage{naaclhlt2018}
\usepackage{times}
\usepackage{amsmath}
\usepackage{url}
\usepackage{tikz}
\usetikzlibrary{shapes,backgrounds,shadows,positioning,fit,matrix,shapes.geometric}
\usepackage[english]{babel}
\usepackage{multirow}
\usepackage{latexsym}
\usepackage{float}
\usepackage{pgfplots}

\aclfinalcopy % Uncomment this line for the final submission
\def\aclpaperid{302} %  Enter the acl Paper ID here

\title{Cross-lingual Abstract Meaning Representation Parsing}

\author{Marco Damonte
  \quad Shay B. Cohen
  \\ \normalsize{School of
    Informatics, University of Edinburgh}\\ \normalsize{10 Crichton Street,
    Edinburgh EH8 9AB, UK}
  \\ \tt{m.damonte@sms.ed.ac.uk} \\
  \tt{scohen@inf.ed.ac.uk}}

\date{}

\begin{document}
\maketitle
\begin{abstract}
  Abstract Meaning Representation (AMR) research has mostly focused on English. We show that it is possible to use AMR annotations for English as a semantic representation for sentences written in other languages. We exploit an AMR parser for English and parallel corpora to learn AMR parsers for Italian, Spanish, German and Chinese. Qualitative analysis show that the new parsers overcome structural differences between the languages. We further propose a method to evaluate the parsers that does not require gold standard data in the target languages. This method highly correlates with the gold standard evaluation, obtaining a (Pearson) correlation of 0.95.
\end{abstract}

\section{Introduction}

Abstract Meaning Representation (AMR) parsing is the process of converting natural language sentences into their corresponding AMR representations \cite{Banarescu13abstractmeaning}. An AMR is a graph with nodes representing the concepts of the sentence and edges representing the semantic relations between them. 
%Most NLP datasets exist only for the English language and AMR parsing is no exception:
Most available AMR datasets large enough to train statistical models consist of pairs of English sentences and AMR graphs.

\begin{figure}
 \centering
  \begin{tikzpicture}
    \draw (0,8) node(z) [] {\emph{This is the sovereignty of each country}};
    \draw (0,6.8) node(s) [ellipse,draw] {sovereignty};
    \draw (2,5.5) node(c) [ellipse,draw] {country};
    \draw (-2,5.5) node(t) [ellipse,draw] {this};
    \draw (2,4) node(e) [ellipse,draw] {each};
    \draw (0,3) node(z) [] {\emph{Questa \`e la sovranit\`a di ogni paese}};
  
     \draw [->] (s) -- node[right=0.1cm]{\small{\emph{:poss}}} (c);
     \draw [->] (s) -- node[right]{\small{\emph{:domain}}} (t);
     \draw [->] (c) -- node[left]{\small{\emph{:mod}}} (e);
     
     \draw [->,dashed] (-2.7,7.8) -- (t);
     \draw [->,dashed] (-2.4,3.2) -- (t);
     \draw [->,dashed] (1.5,3.2) -- (e);
     \draw [->,dashed] (1.5,7.8) to[out=-20,in=20] (e);
     \draw [->,dashed] (0,7.8) -- (s);
     \draw [->,dashed] (2,7.8) -- (c);
     \draw [->,dashed] (0,3.2) to[out=80,in=-80] (s);
     \draw [->,dashed] (2.5,3.2) to[out=-0,in=-10] (c);
  \end{tikzpicture}
 \caption{AMR alignments for a English sentence and its Italian translation.}
 \label{fig:al}
\end{figure}
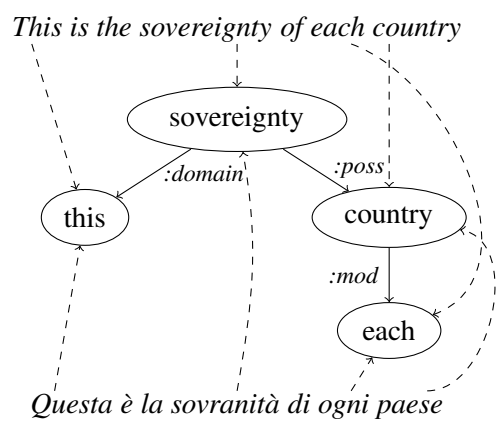

The cross-lingual properties of AMR across languages has been the subject of preliminary discussions.
The AMR guidelines state that AMR is not an interlingua \cite{Banarescu13abstractmeaning} and \newcite{bojar2014comparing} categorizes different kinds of divergences in the annotation between English AMRs and Czech AMRs. \newcite{xue2014not} show that structurally aligning English AMRs with Czech and Chinese AMRs is not always possible but that refined annotation guidelines suffice to resolve some of these cases.
% Building upon this findings, we explore whether AMR can serve as an interlingua, without being one: we ask the question of whether it is possible to maintain the same AMR annotation as a semantic representation for sentences written in other languages, as in Figure~\ref{fig:al}.
We extend this line of research by exploring whether divergences among languages can be overcome, i.e., we investigate whether it is possible to maintain the AMR annotated for English as a semantic representation for sentences written in other languages, as in Figure~\ref{fig:al}.
% it is possible to train statistical models that can retrieve the AMR annotated for English given that the input sentence is written in another language.

We implement AMR parsers for Italian, Spanish, German and Chinese using annotation projection, where existing annotations are projected from a source language (English) to a target language through a parallel corpus \cite[e.g.,][]{yarowsky2001inducing,hwa2005bootstrapping,pado2009cross,evangcross}.
% \footnote{The code and the models for all languages we experimented with are available at \url{https://github.com/mdtux89/amr-eager-multilingual}}
By evaluating the parsers and manually analyzing their output, we show that the parsers are able to recover the AMR structures even when there exist structural differences between the languages, i.e., although AMR is not an interlingua it can act as one. This method also provides a quick way to prototype multilingual AMR parsers, assuming that Part-of-speech (POS) taggers, Named Entity Recognition (NER) taggers and dependency parsers are available for the target languages. We also propose an alternative approach, where Machine Translation (MT) is used to translate the input sentences into English so that an available English AMR parser can be employed. This method is an even quicker solution which only requires translation models between the target languages and English.
% starting from the target parser we learn a new English parser, which we can evaluate against the gold standard. 

Due to the lack of gold standard in the target languages, we exploit the English data to evaluate the parsers for the target languages. (Henceforth, we will use the term target parser to indicate a parser for a target language.)
We achieve this by first learning the target parser from the gold standard English parser, and then inverting this process to learn a new English parser from the target parser. We then evaluate the resulting English parser against the gold standard. We call this ``full-cycle'' evaluation. 

Similarly to \newcite{evangcross}, we also directly evaluate the target parser on ``silver'' data, obtained by parsing the English side of a parallel corpus. 

In order to assess the reliability of these evaluation methods, we collected gold standard datasets for Italian, Spanish, German and Chinese by acquiring professional translations of the AMR gold standard data to these languages.
% \footnote{The translations will be made available upon publication.}
We hypothesize that the full-cycle score can be used as a more reliable proxy than the silver score for evaluating the target parser. We provide evidence to this claim by comparing the three evaluation procedures (silver, full-cycle, and gold) across languages and parsers.

Our main contributions are:
\begin{itemize}
  \item We provide evidence that AMR annotations can be successfully shared across languages.
  \item We propose two ways to rapidly implement non-English AMR parsers.
  \item We propose a novel method to evaluate non-English AMR parsers when gold annotations in the target languages are missing. This method highly correlates with gold standard evaluation, obtaining a Pearson correlation coefficient of 0.95.
  \item We release human translations of an AMR dataset (LDC2015E86) to Italian, Spanish, German and Chinese.
\end{itemize}

\section{Cross-lingual AMR parsing}

AMR is a semantic representation heavily biased towards English, where labels for nodes and edges are either English words or Propbank frames \cite{kingsbury2002treebank}. The goal of AMR is to abstract away from the syntactic realization of the original sentences while maintaining its underlying meaning. 
As a consequence, different phrasings of one sentence are expected to provide identical AMR representations. 
This canonicalization does not always hold across languages: two sentences that express the same meaning in two different languages are not guaranteed to produce identical AMR structures \cite{bojar2014comparing,xue2014not}.  However, \newcite{xue2014not} show that in many cases the unlabeled AMRs are in fact shared across languages. We are encouraged by this finding and argue that it should be possible to develop algorithms that account for some of these differences when they arise. We therefore introduce a new problem, which we call cross-lingual AMR parsing: given a sentence in any language, the goal is to recover the AMR graph that was originally devised for its English translation. 
This task is harder than traditional AMR parsing as it requires to recover English labels as well as to deal with structural differences between languages, usually referred as translation divergence. We propose two initial solutions to this problem: by annotation projection and by machine translation.

\subsection{Method 1: Annotation Projection} 
\label{sec:annotation_projection}

AMR is not grounded in the input sentence, therefore there is no need to change the AMR annotation when projecting to another language. We think of English labels for the graph nodes as ones from an independent language, which incidentally looks similar to English. However, in order to train state-of-the-art AMR parsers, we also need to project the alignments between AMR nodes and words in the sentence (henceforth called AMR alignments). We use word alignments, similarly to other annotation projection work, to project the AMR alignments to the target languages.

Our approach depends on an underlying assumption that we make: if a source word is word-aligned to a target word and it is AMR aligned with an AMR node, then the target word is also aligned to that AMR node. More formally, let $S = s_1 \dots s_{\vert s \vert}$ be the source language sentence and $T = t_1 \dots t_{\vert t \vert}$ be the target language sentence; $A_s(\cdot)$ be the AMR alignment mapping word tokens in $S$ to the set of AMR nodes that are triggered by it; $A_t(\cdot)$ be the same function for $T$; $v$ be a node in the AMR graph; and finally, $W(\cdot)$ be an alignment that maps a word in $S$ to a subset of words in $T$.
%$a_i = j$ indicating that $e_i$ is word aligned with $f_j$; $V_a$ a node in the AMR graph. 
Then, the AMR projection assumption is:
\begin{equation*}
\forall i,j,v \;\; t_j \in W(s_i) \wedge v \in A_s(s_i) \Rightarrow v \in A_t(t_j)
\end{equation*}

% An example is shown in Figure~\ref{fig:al} where both an English sentence and its Italian translation are aligned to the same AMR graph. 
% This is a particularly simple case where all word alignments are 1-to-1 and there are no complications due to translational divergence (i.e., structural differences between languages). 
In the example of Figure~\ref{fig:al}, \emph{Questa} is word-aligned with \emph{This} and therefore AMR-aligned with the node \emph{this}, and the same logic applies to the other aligned words. The words \emph{is}, \emph{the} and \emph{of} do not generate any AMR nodes, so we ignore their word alignments.
We apply this method to project existing AMR annotations to other languages, which are then used to train the target parsers. 
% As there are no parallel corpora with gold AMR annotations available, we use standard MT parallel corpora, which we annotate with AMRs obtained with an available AMR parser for English.

\subsection{Method 2: Machine Translation}
\label{sec:mt}
We invoke an MT system to translate the sentence into English so that we can use an available English parser to obtain its AMR graph. Naturally, the quality of the output graph depends on the quality of the translations. 
% Since the reference is the gold translation with the gold AMR
If the automatic translation is close to the reference translation, then the predicted AMR graph will be close to the reference AMR graph. It is therefore evident that this method is not informative in terms of the cross-lingual properties of AMR. 
% Arguably, this is a workaround rather than a solution to the problem. 
However, its simplicity makes it a compelling engineering solution for parsing other languages.

% We experimented with both phrase-based and neural models as well as with Google Translate\footnote{\url{https://translate.google.com/toolkit}}. 

\subsection{Evaluation}

We now turn to the problem of evaluation. Let us assume that we trained a parser for a target language, for example using the annotation projection method discussed in Section \ref{sec:annotation_projection}. In line with rapid development of new parsers, we assume that the only gold AMR dataset available is the one released for English. %, which provides English sentences with their respective AMR graphs. 
\paragraph{SILVER} We can generate a silver test set by running an automatic (English) AMR parser on the English side of a parallel corpus and use the output AMRs as references. However, the silver test set is affected by mistakes made by the English AMR parser, therefore it may not be reliable.
\paragraph{FULL-CYCLE} In order to perform the evaluation on a gold test set, we propose full-cycle evaluation: after learning the target parser from the English parser, we invert this process to learn a new English parser from the target parser, in the same way that we learned the target parser from the English parser. The resulting English parser is then evaluated against the (English) AMR gold standard. We hypothesize that the score of the new English parser can be used as a proxy to the score of the target parser. 
\paragraph{GOLD} To show whether the evaluation methods proposed can be used reliably, we also generated gold test AMR datasets for four target languages (Italian, Spanish, German and Chinese). In order to do so, we collected professional translations for the English sentences in the AMR test set.\footnote{These datasets are currently available upon request from the authors.} We were then able to create pairs of human-produced sentences with human-produced AMR graphs. 

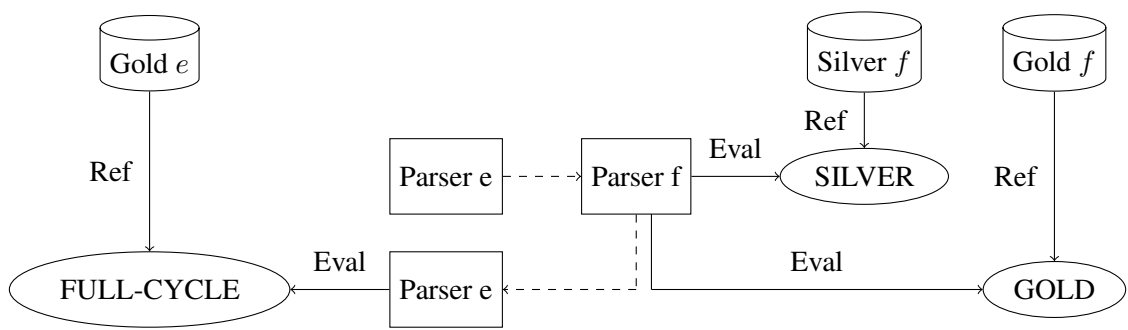
\begin{figure*}[!t]
 \centering
  \begin{tikzpicture}[
    database/.style={
        cylinder,
        shape border rotate=90,
        aspect=0.25,
        draw
     }
  ]
  \draw (-3.9,0) node(gold)[database, minimum width=1cm, minimum height=1cm] {Gold $e$};
  % \draw (1,0) node(par)[database, minimum width=1cm, minimum height=1cm] {Parallel $e \rightarrow f$};
  \draw (5.5,0) node(silv)[database, minimum width=1cm, minimum height=1cm] {Silver $f$};
  \draw (8,0) node(gold2)[database, minimum width=1cm, minimum height=1cm] {Gold $f$};
  % \draw [->] (par) -- (silv) node [midway, above=0.1cm, fill=white] {Project};

  \draw (0,-1.5) node(pe) [rectangle,draw,minimum size=1cm] {Parser e};
  \draw (2.5,-1.5) node(pf) [rectangle,draw,minimum size=1cm] {Parser f};
  \draw (0,-3) node(pe2) [rectangle,draw,minimum size=1cm] {Parser e};
  \draw [-,dashed] (pf) -- (2.5,-3);
  \draw [->,dashed] (2.5,-3) -- (pe2);
  \draw [->,dashed] (pe) -- (pf);

  \draw (-3.9,-3) node(full) [ellipse,draw,minimum size=1cm] {FULL-CYCLE};
  \draw [->] (gold) -- (full) node [midway, left=0.1cm, fill=white] {Ref};
  \draw [->] (pe2) -- (full) node [midway, above=0.1cm, fill=white] {Eval};

  \draw (5.5,-1.5) node(silver) [ellipse,draw] {SILVER};
  \draw [->] (silv) -- (silver) node [midway, left=0.1cm, fill=white] {Ref};
  \draw [->] (pf) -- (silver) node [midway, above=0.1cm, fill=white] {Eval};

  \draw (8,-3) node(goldeval) [ellipse,draw] {GOLD};
  \draw [->] (gold2) -- (goldeval) node [midway, left=0.1cm, fill=white] {Ref};
  \draw [-] (2.7,-2) -- (2.7,-3);
  \draw [->] (2.7,-3) -- (goldeval) node [midway, above=0.1cm, fill=white] {Eval};
  \end{tikzpicture}
 \caption{Description of SILVER, FULL-CYCLE and GOLD evaluations. $e$ stands for English and $f$ stands for the target (foreign) language. Dashed lines represent the process of transferring learning across languages (e.g. with annotation projection). SILVER uses a parsed parallel corpus as reference (``Ref''), FULL-CYCLE uses the English gold standard (Gold $e$) and GOLD uses the target language gold standard we collected (Silver $f$).}
 \label{fig:eval_diagram}
\end{figure*}

A diagram summarizing the different evaluation stages is shown in Figure~\ref{fig:eval_diagram}.
In the case of MT-based systems, the full-cycle corresponds to first translating from English to the target language and then back to English (back-translation), and only then parsing the sentences with the English AMR parser. At the end of this process, a noisy version of the original sentence will be returned and its parsed graph will be a noisy version of the graph parsed from the original sentence. 

\section{Experiments}
\label{sec:experiments}

We run experiments on four languages: Italian, Spanish, German and Chinese. 
We use Europarl \cite{koehn2005europarl} as the parallel corpus for Italian, Spanish and German, containing around 1.9M sentences for each language pair. For Chinese, we use the first 2M sentences from the United Nations Parallel Corpus \cite{ziemski2016united}.
% For training the AMR parsers we extract from each parallel corpus 20,000 sentences for training, 2,000 for development and 2,000 for testing. We collect two such datasets for each language, in order to have non-overlapping datasets for the two stages of the full-cycle process. The remaining sentences are used to train the word alignment models (for annotation projection) and the translation models (for MT). 
For each target language we extract two parallel datasets of 20,000/2,000/2,000 (train/dev/test) sentences for the two step of the annotation projection (English $\rightarrow$ target and target $\rightarrow$ English). These are used to train the AMR parsers. The projection approach also requires training the word alignments, for which we use all the remaining sentences from the parallel corpora (Europarl for Spanish/German/Italian and UN Parallel Corpus for Chinese). These are also the sentences we use to train the MT models.
The gold AMR dataset is LDC2015E86, containing 16,833 training sentences, 1,368 development sentences, and 1,371 testing sentences.

Word alignments were generated using fast\_align \cite{dyer2013simple}, while AMR alignments were generated with {JAMR} \cite{carbonell2014discriminative}. AMREager \cite{damonte2016incremental} was chosen as the pre-existing English AMR parser. AMREager is an open-source AMR parser that needs only minor modifications for re-use with other languages. %\footnote{Available from \url{http://cohort.inf.ed.ac.uk/amreager.html}.}
Our multilingual adaptation of AMREager is available at \url{http://www.github.com/mdtux89/amr-eager-multilingual}.\footnote{A demo is available at \url{http://cohort.inf.ed.ac.uk/amreager.html}.}
It requires tokenization, POS tagging, NER tagging and dependency parsing, which for English, German and Chinese are provided by CoreNLP \cite{corenlp}. We use Freeling \cite{carreras04} for Spanish, as CoreNLP does not provide dependency parsing for this language. Italian is not supported in CoreNLP: we use Tint \cite{aprosio2016italy}, a CoreNLP-compatible NLP pipeline for Italian.

In order to experiment with the approach of Section~\ref{sec:mt}, we experimented with translations from Google Translate.\footnote{\url{https://translate.google.com/toolkit}.} As Google Translate has access to a much larger training corpus, we also trained baseline MT models using Moses \cite{koehn2007moses} and Nematus \cite{sennrich-EtAl:2017:EACLDemo}, with the same training data we use for the projection method and default hyper-parameters. 

% AMR parsers are evaluated with Smatch \cite{cai2013smatch}. 
Smatch \cite{cai2013smatch} is used to evaluate AMR parsers. It looks for the best alignment between the predicted AMR and the reference AMR and it then computes precision, recall and $F_1$ of their edges.
The original English parser achieves 65\% Smatch score on the test split of LDC2015E86. Full-cycle and gold evaluations use the same dataset, while silver evaluation is performed on the split of the parallel corpora we reserved for testing. Results are shown in Table~\ref{tab:results}. The Google Translate system outperforms all other systems, but is not directly comparable to them, as it has the unfair advantage of being trained on a much larger dataset. Due to noisy {JAMR} alignments and silver training data involved in the annotation projection approach, the MT-based systems give in general better parsing results. The BLEU scores of all translation systems are shown in Table~\ref{tab:bleu}.

There are several sources of noise in the annotation projection method, which affect the parsing results: 1) the parsers are trained on silver data obtained by an automatic parser for English; 2) the projection uses noisy word alignments; %\cite{pado2009cross}
3) the AMR alignments on the source side are also noisy; 4) translation divergences exist between the languages, making it sometimes difficult to project the annotation without loss of information.
% For Italian, Spanish and German, the gap between the projection-based parser and the original English parser is around 20\%. This result is comparable to the 15\% gap reported for annotation projection work on Combinatory Categorial Grammar (CCG) parsing \cite{evangcross}.

\begin{table}[t!]
\begin{center}
\begin{tabular}{|ll|c|c|c|}
\hline & \bf System & \bf Silver & \bf Gold & \bf Cycle \\
\multirow{4}{*}{IT} & Projection & 45 & 43 & 45 \\
& Moses & 51 & 52 & 51 \\
& Nematus & 49 & 43 & 41 \\
& GT & 52 & 58 & 59 \\
\hline % \bf Spanish & \bf Silver & \bf Gold & \bf Full-cycle \\
\multirow{4}{*}{ES} & Projection& 44 & 42 & 44 \\
& Moses & 53 & 53 & 51\\
& Nematus & 51 & 43 & 42 \\ 
& GT & 56 & 60 & 60 \\
\hline % \bf German & \bf Silver & \bf Gold & \bf Full-cycle \\
% \multirow{4}{*}{DE} & Projection & 50 & 49 & 49\\
% & Moses & 50 & 50 & 51 \\
\multirow{4}{*}{DE} & Projection & 45 & 39 & 43\\
& Moses & 50 & 49 & 49 \\
& Nematus & 47 & 38 & 39 \\
& GT & 54 & 57 & 59 \\
\hline % \bf Chinese & \bf Silver & \bf Gold & \bf Full-cycle \\
\multirow{4}{*}{ZH} & Projection & 45 & 35 & 32\\
& Moses & 57 & 42 & 48 \\
& Nematus & 57 & 39 & 40 \\
& GT & 64 & 50 & 55 \\
\hline
\end{tabular}
\end{center}
\caption{Silver, gold and full-cycle Smatch scores for projection-based and MT-based systems.}
\label{tab:results}
\end{table}

% moses
% LANGUAGE de
% silver (deeuroparl)
% Smatch -> P: 0.51, R: 0.49, F: 0.50
% gold
% Smatch -> P: 0.52, R: 0.46, F: 0.49
% full-cycle
% Smatch -> P: 0.53, R: 0.46, F: 0.49

% proj
% LANGUAGE de
% silver (deeuroparl)
% 45
% gold
% 39
% full-cycle
% 43

%missing: nematus, gtranslate, projection

\begin{table}[t!]
\begin{center}
\begin{tabular}{|l|c|c|c|}
\hline \bf Model & \bf Moses & \bf Nematus & \bf GT \\
EN-IT & 23.83 & 21.27 & 61.31 \\
IT-EN & 23.74 & 19.77 & 42.20 \\
EN-ES & 29.00 & 26.14 & 78.14 \\
ES-EN & 27.66 & 21.63 & 50.78 \\
EN-DE & 15.47 & 15.74 & 63.48 \\ 
DE-EN & 21.50 & 14.96 & 41.78 \\ 
% EN-ZH & 4.38 & 0.68 & 26.60 \\
% ZH-EN & 3.75 & 0.87 & 22.21 \\
EN-ZH & 9.19 & 8.67 & 26.75 \\
ZH-EN & 10.81 & 10.37 & 22.21 \\
\hline
\end{tabular}
\end{center}
\caption{BLEU scores for Moses, Nematus and Google Translate (GT) on the (out-of-domain) LDC2015E86 test set}
\label{tab:bleu}
\end{table}

\section{Qualitative Analysis}

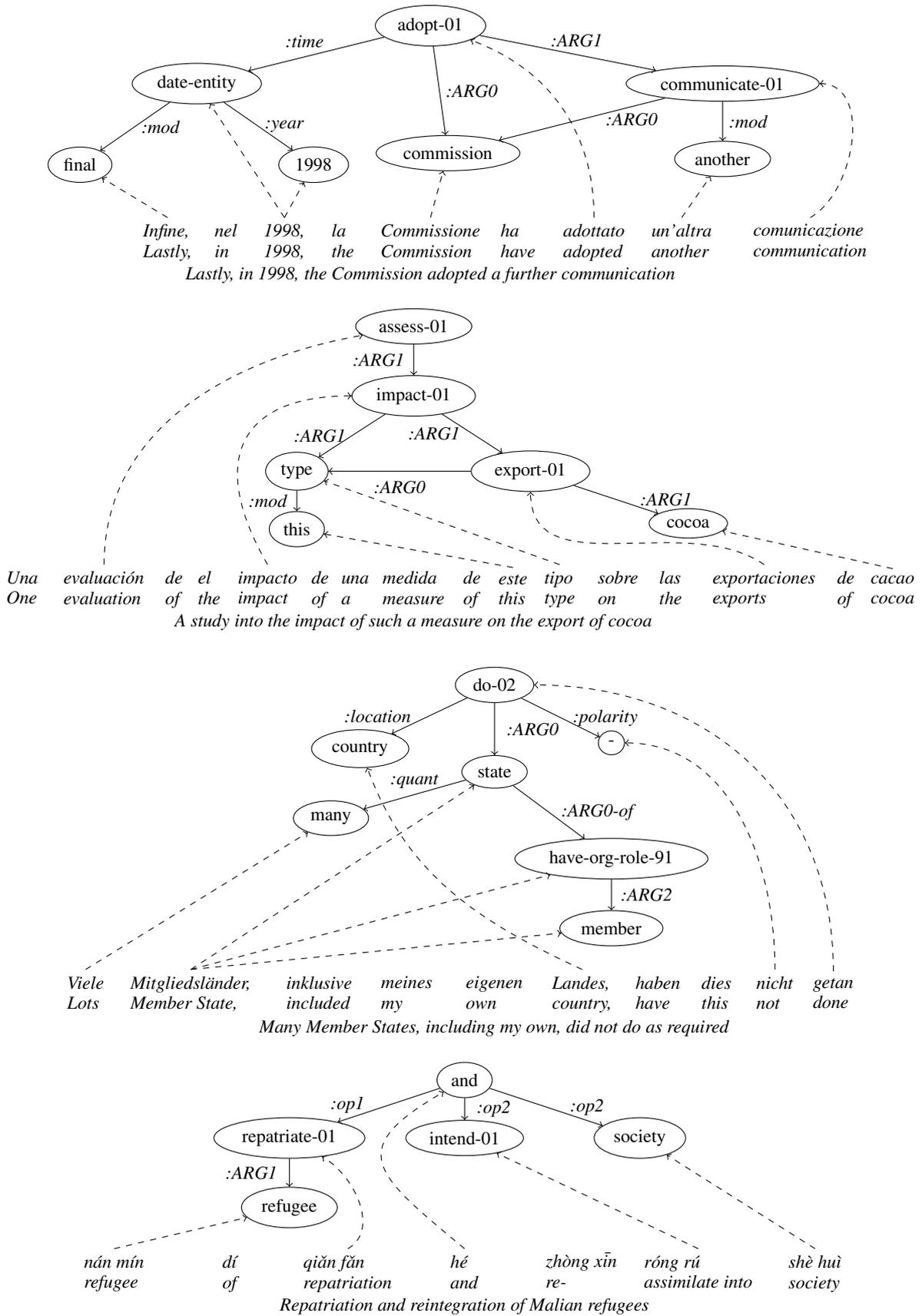
\begin{figure*}[t!]
\small{
\centering
   \begin{tikzpicture}
    \draw (0,8) node(a) [ellipse,draw] {adopt-01};
    \draw (-4,7) node(d) [ellipse,draw] {date-entity};
    \draw (0.3,5.8) node(c) [ellipse,draw] {commission};
    \draw (5,7) node(c2) [ellipse,draw] {communicate-01};
    \draw (-2,5.6) node(t) [ellipse,draw] {1998};
    \draw (-6,5.6) node(f) [ellipse,draw] {final};
    \draw (5,5.7) node(a2) [ellipse,draw] {another};

    \draw (-4.5,4.8) node(z1) [label={[align=left]below:\emph{Infine,}\\\emph{Lastly,}}] {};
    \draw (-3.5,4.8) node(z2) [label={[align=left]below:\emph{nel}\\\emph{in}}] {};
    \draw (-2.5,4.8) node(z3) [label={[align=left]below:\emph{1998,}\\\emph{1998,}}] {};
    \draw (-1.5,4.8) node(z4) [label={[align=left]below:\emph{la}\\\emph{the}}] {};
    \draw (0,4.8) node(z5) [label={[align=left]below:\emph{Commissione}\\\emph{Commission}}] {};
    \draw (1.5,4.8) node(z6) [label={[align=left]below:\emph{ha}\\\emph{have}}] {};
    \draw (2.8,4.8) node(z7) [label={[align=left]below:\emph{adottato}\\\emph{adopted}}] {};
    \draw (4.3,4.8) node(z8) [label={[align=left]below:\emph{un'altra}\\\emph{another}}] {};    
    \draw (6.5,4.8) node(z9) [label={[align=left]below:\emph{comunicazione}\\\emph{communication}}] {};
    \draw (0,3.7) node() {\emph{Lastly, in 1998, the Commission adopted a further communication}};

    \draw [->] (a) -- node[right]{\small{\emph{:ARG0}}} (c);
    \draw [->] (a) -- node[left=0.2cm,above]{\small{\emph{:time}}} (d);
    \draw [->] (a) -- node[right=0.15cm,above]{\small{\emph{:ARG1}}} (c2);
    \draw [->] (d) -- node[right]{\small{\emph{:year}}} (t);
    \draw [->] (d) -- node[right]{\small{\emph{:mod}}} (f);
    \draw [->] (c2) -- node[right]{\small{\emph{:mod}}} (a2);
    \draw [->] (c2) -- node[below,right=0.3cm]{\small{\emph{:ARG0}}} (c);

    \draw [->,dashed] (z7.south) to[out=90,in=-20] (a);
    \draw [->,dashed] (z9.south) to[out=30,in=0] (c2);
    \draw [->,dashed] (z8.south) -- (a2);
    \draw [->,dashed] (z3.south) -- (t);
    \draw [->,dashed] (z3.south) -- (d);
    \draw [->,dashed] (z1.south) -- (f);
    \draw [->,dashed] (z5.south) -- (c);
  \end{tikzpicture}
  \vspace{0.3cm}
  
  \begin{tikzpicture}
      \draw (0,8) node(a) [ellipse,draw] {assess-01};
      \draw (0,6.8) node(i) [ellipse,draw] {impact-01};
      \draw (-2,5.5) node(t) [ellipse,draw] {type};
      \draw (2,5.5) node(e) [ellipse,draw] {export-01};
      \draw (-2,4.5) node(t2) [ellipse,draw] {this};
      \draw (4.7,4.6) node(c) [ellipse,draw] {cocoa};
      \draw (-6.7,4) node(z1) [label={[align=left]below:\emph{Una}\\\emph{One}}] {};
      \draw (-5.3,4) node(z2) [label={[align=left]below:\emph{evaluaci\'on}\\\emph{evaluation}}] {};
      \draw (-4.1,4) node(z3) [label={[align=left]below:\emph{de}\\\emph{of}}] {};
      \draw (-3.5,4) node(z4) [label={[align=left]below:\emph{el}\\\emph{the}}] {};
      \draw (-2.5,4) node(z5) [label={[align=left]below:\emph{impacto}\\\emph{impact}}] {};
      \draw (-1.6,4) node(z6) [label={[align=left]below:\emph{de}\\\emph{of}}] {};
      \draw (-1,3.92) node(z7) [label={[align=left]below:\emph{una}\\\emph{a}}] {};
      \draw (0,4) node(z8) [label={[align=left]below:\emph{medida}\\\emph{measure}}] {};
      \draw (1,4) node(z9) [label={[align=left]below:\emph{de}\\\emph{of}}] {};
      \draw (1.7,3.95) node(z10) [label={[align=left]below:\emph{este}\\\emph{this}}] {};
      \draw (2.5,4) node(z11) [label={[align=left]below:\emph{tipo}\\\emph{type}}] {};
      \draw (3.5,4) node(z12) [label={[align=left]below:\emph{sobre}\\\emph{on}}] {};
      \draw (4.4,4) node(z13) [label={[align=left]below:\emph{las}\\\emph{the}}] {};
      \draw (6,4) node(z14) [label={[align=left]below:\emph{exportaciones}\\\emph{exports}}] {};
      \draw (7.4,4) node(z15) [label={[align=left]below:\emph{de}\\\emph{of}}] {};
      \draw (8.2,3.93) node(z16) [label={[align=left]below:\emph{cacao}\\\emph{cocoa}}] {};
      \draw (0,2.9) node() {\emph{A study into the impact of such a measure on the export of cocoa}};

      \draw [->] (a) -- node[left=0.05cm]{\small{\emph{:ARG1}}} (i);
      \draw [->] (i) -- node[left]{\small{\emph{:ARG1}}} (t);
      \draw [->] (i) -- node[left=0.1cm]{\small{\emph{:ARG1}}} (e);
      \draw [->] (t) -- node[left=0.05cm]{\small{\emph{:mod}}} (t2);
      \draw [->] (e) -- node[right=0.3cm]{\small{\emph{:ARG1}}} (c);
      \draw [->] (e) -- node[below=0.05cm]{\small{\emph{:ARG0}}} (t);
      
      \draw [->,dashed] (z2.south) to[out=90,in=190] (a);
      \draw [->,dashed] (z5.south) to[out=120,distance=2cm,in=180] (i);
      \draw [->,dashed] (z11.south) -- (t);
      \draw [->,dashed] (z14.south) to[out=145,in=-90] (e);
      \draw [->,dashed] (z10.south) -- (t2);
      \draw [->,dashed] (z16.south) -- (c);
  \end{tikzpicture}
    \vspace{0.2cm}

  \begin{tikzpicture}
      \draw (0,8.5) node(dummy) [] {};
      \draw (0,8) node(d) [ellipse,draw] {do-02};
      \draw (-2.3,6.9) node(c) [ellipse,draw] {country};
      \draw (0,6.5) node(s) [ellipse,draw] {state};
      \draw (2,7) node(p) [ellipse,draw] {-};
      \draw (-2.8,5.7) node(m) [ellipse,draw] {many};
      \draw (2,5) node(h) [ellipse,draw] {have-org-role-91};
      \draw (2,3.8) node(m2) [ellipse,draw] {member};
      \draw (-7,3.2) node(z1) [label={[align=left]below:\emph{Viele}\\\emph{Lots}}] {};
      \draw (-5.2,3.2) node(z2) [label={[align=left]below:\emph{Mitgliedsl\"ander,}\\\emph{Member State,}}] {};
      \draw (-3,3.2) node(z3) [label={[align=left]below:\emph{inklusive}\\\emph{included}}] {};
      \draw (-1.5,3.2) node(z4) [label={[align=left]below:\emph{meines}\\\emph{my}}] {};
      \draw (0,3.2) node(z5) [label={[align=left]below:\emph{eigenen}\\\emph{own}}] {};
      \draw (1.5,3.2) node(z6) [label={[align=left]below:\emph{Landes,}\\\emph{country,}}] {};
      \draw (2.8,3.2) node(z7) [label={[align=left]below:\emph{haben}\\\emph{have}}] {};
      \draw (3.8,3.2) node(z8) [label={[align=left]below:\emph{dies}\\\emph{this}}] {};
      \draw (4.8,3.2) node(z9) [label={[align=left]below:\emph{nicht}\\\emph{not}}] {};
      \draw (5.8,3.17) node(z10) [label={[align=left]below:\emph{getan}\\\emph{done}}] {};
      \draw (0,2.07) node() {\emph{Many Member States, including my own, did not do as required}};
      
      \draw [->] (d) -- node[left=0.2cm]{\small{\emph{:location}}} (c);
      \draw [->] (d) -- node[right=0.1cm]{\small{\emph{:polarity}}} (p);
      \draw [->] (d) -- node[right=0.1cm]{\small{\emph{:ARG0}}} (s);
      \draw [->] (s) -- node[right=0.1cm]{\small{\emph{:ARG0-of}}} (h);
      \draw [->] (s) -- node[above]{\small{\emph{:quant}}} (m);
      \draw [->] (h) -- node[right=0.05cm]{\small{\emph{:ARG2}}} (m2);
      
      \draw [->,dashed] (z1.south) -- (m);
      \draw [->,dashed] (z2.south) -- (s);
      \draw [->,dashed] (z2.south) -- (h);
      \draw [->,dashed] (z2.south) -- (m2);
      \draw [->,dashed] (z6.south) to[out=155,in=-70] (c);
      \draw [->,dashed] (z9.south) to[out=90,in=0] (p);
      \draw [->,dashed] (z10.south) to[out=90,in=0,distance=3cm] (d);
      
  \end{tikzpicture}
    \vspace{0.3cm}

  \begin{tikzpicture}

  \draw (0,8) node(and) [ellipse,draw] {and};
  \draw (0,7) node(intend) [ellipse,draw] {intend-01};
  \draw (3,7) node(society) [ellipse,draw] {society};
  \draw (-3,7) node(repat) [ellipse,draw] {repatriate-01};
  \draw (-3,5.8) node(refugee) [ellipse,draw] {refugee};
  % \draw (-3.5,5) node(c) [ellipse,draw] {country};
  % \draw (-2.5,4.2) node(n) [ellipse,draw] {name};
  % \draw (-2.5,3.5) node(f) [ellipse,draw] {Mali};
  % \draw (-0.5,4.8) node(mali) [ellipse,draw] {Mali};
  
  \draw [->] (and) -- node[right=0.1cm]{\small{\emph{:op2}}} (intend); 
  \draw [->] (and) -- node[right=0.3cm]{\small{\emph{:op2}}} (society);
  \draw [->] (and) -- node[left=0.3cm]{\small{\emph{:op1}}} (repat);
  \draw [->] (repat) -- node[left=0.1cm]{\small{\emph{:ARG1}}} (refugee);  
  % \draw [->] (refugee) -- node[left=0.1cm]{\small{\emph{:country}}} (mali); 

  \draw (-6,5.2) node(z1) [label={[align=left]below:\emph{n\'an m\'in}\\\emph{refugee}}] {};
  \draw (-4,5.2) node(z2) [label={[align=left]below:\emph{d\' i}\\\emph{of}}] {};
  \draw (-2,5.2) node(z3) [label={[align=left]below:\emph{qi\v an f\v an}\\\emph{repatriation}}] {};
  \draw (0,5.2) node(z4) [label={[align=left]below:\emph{h\'e}\\\emph{and}}] {};
  \draw (2,5.25) node(z5) [label={[align=left]below:\emph{zh\`ong x\=in}\\\emph{re-}}] {};
  \draw (4,5.2) node(z6) [label={[align=left]below:\emph{r\'ong r\'u}\\\emph{assimilate into}}] {};
  \draw (6,5.2) node(z7) [label={[align=left]below:\emph{sh\`e hu\`i}\\\emph{society}}] {};
  \draw (0,4.1) node() {\emph{Repatriation and reintegration of Malian refugees}};

  \draw [->,dashed] (z1.south) -- (refugee);
  \draw [->,dashed] (z3.south) to[out=30,in=-30] (repat);
  \draw [->,dashed] (z4.south) to[out=140,in=210,distance=2cm] (and);
  \draw [->,dashed] (z6.south) -- (intend);
  \draw [->,dashed] (z7.south) -- (society);
  % \draw [->,dashed] (z6.south) to[out=155,in=-70] (c);
  % \draw [->,dashed] (z9.south) to[out=90,in=0] (p);
  % \draw [->,dashed] (z10.south) to[out=90,in=0,distance=3cm] (d);
% # ::id 0
% # # ::snt 年 6 月 签署 了 一 项 谅解 备忘录 , 概述 此 种 合作 的 框架 .
% # # ::alignments 0-3|0.0+0.0.0 3-4|0 7-8|0.1.0 8-9|0.1 10-11|0.2 13-14|0.2.0.0 15-16|0.2.0
% # (v3 / sign-01
% #     :time (v1 / temporal-quantity
% #             :unit (v2 / month))
% #                 :ARG1 (v5 / memorandum
% #                         :topic (v4 / understand-01))
% #                             :ARG1 (v6 / outline-01
% #                                     :ARG1 (v8 / framework
% #                                                 :ARG0-of (v7 / cooperate-01))))

  \end{tikzpicture}
 % \caption{Parsed AMR graph and alignments (dashed lines) for the Italian translation of \emph{Lastly, in 1998, the Commission adopted a further communication}, the Spanish translation of \emph{A study into the impact of such a measure on the export of cocoa}, the German translation of \emph{Many Member States, including my own, did not do as required} and the Chinese translation of \emph{Repatriation and reintegration of Malian refugees}.}
  \caption{Parsed AMR graph and alignments (dashed lines) for an Italian sentence, a Spanish sentence, a German sentences and a Chinese sentence.}
 \label{fig:examples}
 }
 \end{figure*}

Figure~\ref{fig:examples} shows examples of output parses\footnote{In this section, all parsed graphs were generated with the projection-based system of Section~\ref{sec:annotation_projection}.} for all languages, including the AMR alignments by-product of the parsing process, that we use to discuss the mistakes made by the parsers.

In the Italian example, the only evident error is that \emph{Infine} (\emph{Lastly}) should be ignored. 
In the Spanish example, the word \emph{medida} (\emph{measure}) is wrongly ignored: it should be used to generate a child of the node \emph{impact-01}. Some of the \emph{:ARG} roles are also not correct. 
In the German example, \emph{meines} (\emph{my}) should reflect the fact that the speaker is talking about his own country. Finally, in the Chinese example, there are several mistakes including yet another concept identification mistake: \emph{intend-01} is erroneously triggered.

Most mistakes involve concept identification. In particular, relevant words are often erroneously ignored by the parser. This is directly related to the problem of noisy word alignments in annotation projection: the parser learns what words are likely to trigger a node (or a set of nodes) in the AMR by looking at their AMR alignments (which are induced by the word alignments). If an important word consistently remains unaligned, the parser will erroneously learn to discard it. More accurate alignments are therefore crucial in order to achieve better parsing results. We computed the percentage of words in the training data that are learned to be non-content-bearing in each parser and we found that the Chinese parser, which is our least accurate parser, is the one that most suffer from this, with 33\% non-content-bearing words. On the other hand, in the German parser, which is the highest scoring, only 26\% of the words are non-content-bearing, which is the lowest percentage amongst all parsers.

\subsection{Translational Divergence}

In order to investigate the hypothesis that AMR can be shared across these languages, we now look at translational divergence and discuss how it affects parsing, following the classification used in previous work \cite{dorr2002improved,dorr1994machine}, which identifies classes of divergences for several languages. \newcite{sulem2015conceptual} also follow the same categorization for French.

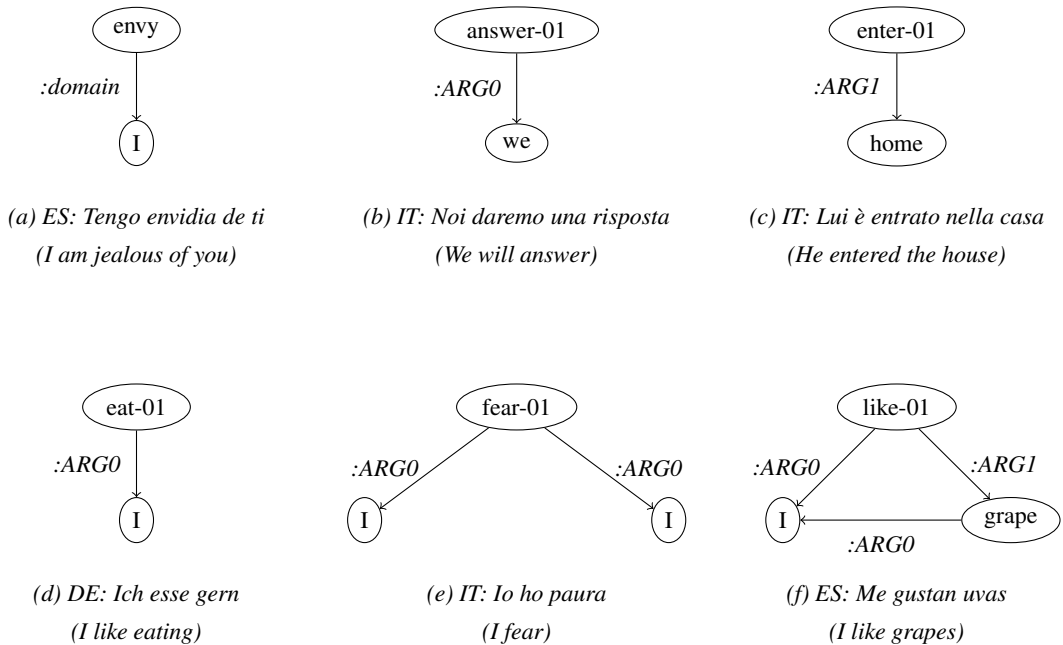
\begin{figure*}[t]
  \centering
  \small
  \begin{tikzpicture}
  \draw (0,8) node(j) [ellipse,draw] {envy};
  \draw (0,6.5) node(i) [ellipse,draw] {I};
  \draw [->] (j) -- node[left=0.1cm]{\small{\emph{:domain}}} (i);
  \draw (0,5.5) node(z1) [] {\emph{(a) ES: Tengo envidia de ti}};
  \draw (0,5) node(z1) [] {\emph{(I am jealous of you)}};

        \draw (5,8) node(a) [ellipse,draw] {answer-01};
  \draw (5,6.5) node(w) [ellipse,draw] {we};
  \draw [->] (a) -- node[left=0.1cm]{\small{\emph{:ARG0}}} (w);
  \draw (5,5.5) node(z2) [] {\emph{(b) IT: Noi daremo una risposta}};
        \draw (5,5) node(z2) [] {\emph{(We will answer)}};
        
        \draw (10,8) node(e) [ellipse,draw] {enter-01};
  \draw (10,6.5) node(h) [ellipse,draw] {home};
  \draw [->] (e) -- node[left=0.1cm]{\small{\emph{:ARG1}}} (h);
  \draw (10,5.5) node(z3) [] {\emph{(c) IT: Lui \`e entrato nella casa}};
        \draw (10,5) node(z3) [] {\emph{(He entered the house)}};
        
  \draw (0,3) node(l) [ellipse,draw] {eat-01};
  \draw (0,1.5) node(i) [ellipse,draw] {I};
  \draw [->] (l) -- node[left=0.1cm]{\small{\emph{:ARG0}}} (i);
        \draw (0,0.5) node(z1) [] {\emph{(d) DE: Ich esse gern}};
        \draw (0,0) node(z3) [] {\emph{(I like eating)}};
        
        \draw (5,3) node(f) [ellipse,draw] {fear-01};
  \draw (3,1.5) node(fi) [ellipse,draw] {I};
  \draw (7,1.5) node(si) [ellipse,draw] {I};
  \draw [->] (f) -- node[left=0.1cm]{\small{\emph{:ARG0}}} (fi);
  \draw [->] (f) -- node[right=0.1cm]{\small{\emph{:ARG0}}} (si);
  \draw (5,0.5) node(z3) [] {\emph{(e) IT: Io ho paura}};
        \draw (5,0) node(z3) [] {\emph{(I fear)}};

  \draw (10,3) node(l) [ellipse,draw] {like-01};
  \draw (8.5,1.5) node(i) [ellipse,draw] {I};
  \draw (11.5,1.5) node(g) [ellipse,draw] {grape};
  \draw [->] (l) -- node[right=0.1cm]{\small{\emph{:ARG1}}} (g);
  \draw [->] (l) -- node[left=0.1cm]{\small{\emph{:ARG0}}} (i);
  \draw [->] (g) -- node[below=0.1cm]{\small{\emph{:ARG0}}} (i);
  \draw (10,0.5) node(z3) [] {\emph{(f) ES: Me gustan uvas}};
  \draw (10,0) node(z3) [] {\emph{(I like grapes)}};
  \end{tikzpicture}
 \caption{Parsing examples in several languages involving common translational divergence phenomena: (a) contains a categorical divergence, (b) and (e) conflational divergences, (c) a structural divergence, (d) an head swapping and (f) a thematic divergence.}
 \label{fig:divergences}
\end{figure*}

Figure~\ref{fig:divergences} shows six sentences displaying these divergences. The aim of this analysis is to assess how the parsers deal with the different kind of translational divergences, regardless of the overall quality of the output. 

\paragraph{Categorical.} This divergence happens when two languages use different POS tags to express the same meaning. For example, the English sentence \emph{I am jealous of you} is translated into Spanish as \emph{Tengo envidia de ti} (\emph{I have jealousy of you}). The English adjective \emph{jealous} is translated in the Spanish noun \emph{envidia}. In Figure~\ref{fig:divergences}a we note that the categorical divergence does not create problems since the parsers correctly recognized that \emph{envidia} (\emph{jealousy}/\emph{envy}) should be used as the predicate, regardless of its POS.

\paragraph{Conflational.} This divergence happens when verbs expressed in a language with a single word can be expressed with more words in another language. Two subtypes are distinguished: \emph{manner} and \emph{light verb}. Manner refers to a manner verb that is mapped to a motion verb plus a manner-bearing word. For example, \emph{We will answer} is translated in the Italian sentence \emph{Noi daremo una riposta} (\emph{We will give an answer}), where \emph{to answer} is translated as \emph{daremo una risposta} (\emph{will give an answer}). Figure~\ref{fig:divergences}b shows that the Italian parser generates a sensible output for this sentence by creating a single node labeled \emph{answer-01} for the expression \emph{dare una riposta}. 

In a light verb conflational divergence, a verb is mapped to a light verb plus an additional meaning unit, such as when \emph{I fear} is translated as \emph{Io ho paura} (\emph{I have fear}) in Italian: \emph{to fear} is mapped to the light verb \emph{ho} (\emph{have}) plus the noun \emph{paura} (\emph{fear}). Figure~\ref{fig:divergences}e shows that also this divergence is dealt properly by the Italian parser: \emph{ho paura} correctly triggers the root \emph{fear-01}. 

\paragraph{Structural.} This divergence happens when verb arguments result in different syntactic configurations, for example, due to an additional PP attachment. When translating \emph{He entered the house} with \emph{Lui \`e entrato nella casa} (\emph{He entered in the house}), the Italian translation has an additional \emph{in} preposition. Also this parsed graph, in Figure~\ref{fig:divergences}c, is structurally correct. The missing node \emph{he} is due to pronoun-dropping, which is frequent in Italian.

\paragraph{Head swapping.} This divergence occurs when the direction of the dependency between two words is inverted. For example, \emph{I like eating}, where \emph{like} is head of \emph{eating}, becomes \emph{Ich esse gern} (\emph{I eat likingly}) in German, where the dependency is inverted. Unlike all other examples, in this case, the German parser does not cope well with this divergence: it is unable to recognize \emph{like-01} as the main concept in the sentence, as shown in Figure~\ref{fig:divergences}d. 

\paragraph{Thematic.} Finally, the parse of Figure~\ref{fig:divergences}f has to deal with a thematic divergence, which happens when the semantic roles of a predicate are inverted. In the sentence \emph{I like grapes}, translated to Spanish as \emph{Me gustan uvas}, \emph{I} is the subject in English while \emph{Me} is the object in Spanish. Even though we note an erroneous reentrant edge between \emph{grape} and \emph{I}, the thematic divergence does not create problems: the parser correctly recognizes the \emph{:ARG0} relationship between \emph{like-01} and \emph{I} and the \emph{:ARG1} relationship between \emph{like-01} and \emph{grape}. In this case, the edge labels are important, as this type of divergence is concerned with the semantic roles.

\section {Discussion}

\begin{figure}[h!]  
\centering 
\begin{tikzpicture}[scale=0.65]
\begin{axis}[
    axis lines=middle,
    x label style={at={(current axis.right of origin)},anchor=north, below=5mm},
    y label style={at={(current axis.above origin)},rotate=90,anchor=south east, left=10mm},    
    xlabel={Gold Smatch},
    ylabel={Silver Smatch},
    xmin=30, xmax=70,
    ymin=30, ymax=70]
\addplot [only marks] table {
43 45
52 51
43 49
58 52
42 44
53 53
43 51
60 56
39 45
49 50
38 47
57 54
35 45
42 57
39 57
50 64
};
\addplot [domain=30:70, samples=2, dashed] {0.332165670719*x + 35.825056666};
\end{axis}
\end{tikzpicture}
% \hspace{1cm}%

\begin{tikzpicture}[scale=0.65]
\begin{axis}[
    axis lines=middle,
    x label style={at={(current axis.right of origin)},anchor=north, below=5mm},
    y label style={at={(current axis.above origin)},rotate=90,anchor=south east, left=10mm},
    xlabel={Gold Smatch},
    ylabel={Full-cycle Smatch},    
    xmin=30, xmax=70,
    ymin=30, ymax=70]
\addplot [only marks] table {
43 45
52 51
43 41
58 59
42 44
53 51
43 42
60 60
39 43
49 49
38 39
57 59
35 32
42 48
39 40
50 55
};
\addplot [domain=30:70, samples=2, dashed] {0.987293083316*x + 1.52757744351};
\end{axis}
\end{tikzpicture}
% \caption{Top: Linear regression for silver evaluation. Bottom: Linear regression for full-cycle evaluation. 
\caption{Linear regression lines for silver and full-cycle.
%The Smatch scores for gold evaluation are on the x-axis and on the silver and full-cycle scores are on the y-axis.
}
\label{fig:regr}
\end{figure}
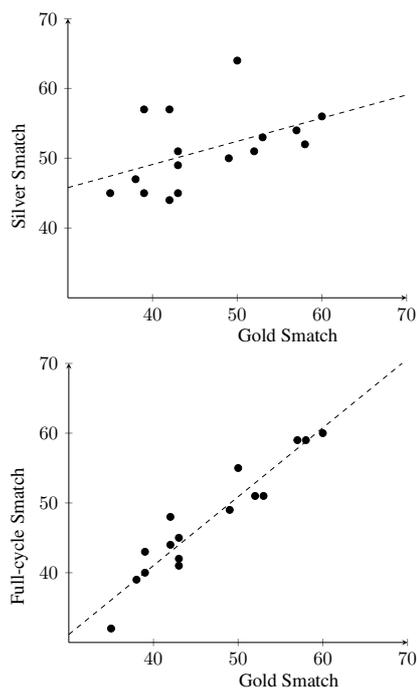

\paragraph{Can AMR be shared across these languages?}
As mentioned in Section~\ref{sec:mt}, the MT-based systems are not helpful in answering this question and we instead focus on the projection-based parsers. Qualitative analysis showed that the parsers are able to overcome translational divergence and that concept identification must be more accurate in order to provide good parsing results. We therefore argue that the suboptimal performance of the parsers in terms of Smatch scores is due to the many sources of noise in the annotation projection approach rather than instability of AMR across languages. We provide strong evidence that cross-lingual AMR parsing is indeed feasible and hope that the release of the gold standard test sets will motivate further work in this direction.
\paragraph{Are silver and full-cycle evaluations reliable?}
We computed the Pearson correlation coefficients for the Smatch scores of Table~\ref{tab:results} to determine how well silver and full-cycle correlate with gold evaluation. Full-cycle correlates better than silver: the Pearson coefficient is 0.95 for full-cycle and 0.47 for silver. Figure~\ref{fig:regr} shows linear regression lines. Unlike silver, full-cycle uses the same dataset as gold evaluation and it does not contain parsing mistakes, which makes it more reliable than silver.
Interestingly, if we ignore the scores obtained for Chinese, the correlation between silver and gold dramatically increases, perhaps indicating that Europarl is more suitable than the UN corpus for this task: the Pearson coefficient becomes 0.97 for full-cycle and 0.87 for silver. 
A good proxy for gold evaluation should rank different systems similarly. We hence computed the Kendall-tau score \cite{kendall1945treatment}, a measure for similarity between permutations, of the rankings extracted from Table 1. The results further confirm that full-cycle approximate gold better than silver does: the score is 0.40 for silver and 0.82 for full-cycle.
Full cycle introduces additional noise but it is not as expensive as gold and is more reliable than silver. 

\section {Related Work}
% Cross-lingual alignments can be inspected by using AMRICA \cite{saphra2015amrica}. The only work that attempts to automatically parse AMR graphs for non-English sentences is \cite{vanderwende2015amr}. Sentences in several languages (French, German, Spanish and Japanese) are parsed into a logical representation, which is then converted to AMR using a small set of rules.

AMR parsing for languages other than English has made only a few steps forward. In previous work \cite{li2016annotating,xue2014not,bojar2014comparing}, nodes of the target graph were labeled with either English words or with words in the target language. We instead use the AMR annotation used for English for the target language as well, without translating any word.
% Cross-lingual approaches of the model transfer type, where parameters are shared across languages \cite{zeman2008cross,cohen-11b,mcdonald2011multi,sogaard2011data}, would also benefit from a monolingual lexicon in the annotations. 
% \newcite{li2016annotating} annotated the Chinese translation of ``The Little Prince'' with AMRs. \newcite{bojar2014comparing} and \newcite{xue2014not} investigated whether AMR can be considered as an interlingua, even though stated otherwise in \newcite{Banarescu13abstractmeaning}. They found that in certain cases AMR graphs are structurally similar across languages. 
% A tool for visualizing cross-lingual alignments between AMR graphs was also implemented \cite{saphra2015amrica}. 
To the best of our knowledge, the only previous work that attempts to automatically parse AMR graphs for non-English sentences is by \newcite{vanderwende2015amr}. Sentences in several languages (French, German, Spanish and Japanese) are parsed into a logical representation, which is then converted to AMR using a small set of rules. A comparison with this work is difficult, as the authors do not report results for the parsers (due to the lack of an annotated corpus) or release their code.

Besides AMR, other semantic parsing frameworks for non-English languages have been investigated \cite{hoffman1992ccg,cinkova2009tectogrammatical,gesmundo2009latent,evangcross}. \newcite{evangcross} is the most closely related to our work as it uses a projection mechanism similar to ours for CCG. A crucial difference is that, in order to project CCG parse trees to the target languages, they only make use of literal translation.
% , which we argue is not as necessary in our case since AMR is expected to abstract away from the different syntactic realizations.
% On assessing stability across languages:
Previous work has also focused on assessing the stability across languages of semantic frameworks such as AMR \cite{xue2014not,bojar2014comparing}, UCCA \cite{sulem2015conceptual} and Propbank \cite{van2010cross}. 
% This work assumes that AMR is sufficiently stable and enforces it by making the AMR graph annotated for English valid for the target language as well. Supporting evidence comes from the preliminary work on Chinese \cite{xue2014not} and investigation of the cross-lingual stability of Propbank, on which AMR is partially based \cite{van2010cross}.
%with the exceptions of idioms, collocations, and multi-word expressions is shown.

%On Cross-lingual:
Cross-lingual techniques can cope with the lack of labeled data on languages when this data is available in at least one language, usually English. The annotation projection method, which we follow in this work, is one way to address this problem. It was introduced for POS tagging, base noun phrase bracketing, NER tagging, and inflectional morphological analysis \cite{yarowsky2001inducing} but it has also been used for dependency parsing \cite{hwa2005bootstrapping}, role labeling \cite{pado2009cross,akbik2015generating} and semantic parsing \cite{evangcross}. Another common thread of cross-lingual work is model transfer, where parameters are shared across languages \cite{zeman2008cross,cohen2009shared,cohen-11b,mcdonald2011multi,sogaard2011data}.

\section{Conclusions}
We introduced the problem of parsing AMR structures, annotated for English, from sentences written in other languages as a way to test the cross-lingual properties of AMR. We provided evidence that AMR can be indeed shared across the languages tested and that it is possible to overcome translational divergences. We further proposed a novel way to evaluate the target parsers that does not require manual annotations of the target language. The full-cycle procedure is not limited to AMR parsing and could be used for other cross-lingual problems in NLP. The results of the projection-based AMR parsers indicate that there is a vast room for improvements, especially in terms of generating better alignments. We encourage further work in this direction by releasing professional translations of the AMR test set into four languages.

\section*{Acknowledgments}
The authors would like to thank the three anonymous reviewers and Sameer Bansal, Gozde Gul Sahin, Sorcha Gilroy, Ida Szubert, Esma Balkir, Nikos Papasarantopoulos, Joana Ribeiro, Shashi Narayan, Toms Bergmanis, Clara Vania, Yang Liu and Adam Lopez for their helpful comments. This research was supported by a grant from Bloomberg and by the H2020 project SUMMA, under grant agreement 688139.
% Ida (for translation idea)
% Joachim (for being awesome)

% Acknowledgments
% Adam, Yang, Kristina.
\bibliography{naaclhlt2018-cross}
\bibliographystyle{acl_natbib}

\end{document}